\documentclass[conference]{IEEEtran}
\IEEEoverridecommandlockouts
\usepackage{amsmath,amssymb,amsfonts}
\usepackage{algorithmic}
\usepackage{graphicx}
\usepackage{textcomp}
\usepackage{xcolor}
\def\BibTeX{{\rm B\kern-.05em{\sc i\kern-.025em b}\kern-.08em
    T\kern-.1667em\lower.7ex\hbox{E}\kern-.125emX}}

\usepackage[
backend=biber,
style=ieee,
]{biblatex} %Imports biblatex package
\usepackage[caption=false, font=footnotesize]{subfig}

\usepackage{booktabs,tabularx}
\usepackage{array}
\usepackage{multirow}

\begin{document}

%\title{Perturbed Report Discrimination Improves Biomedical Vision-language Models}
%{\footnotesize \textsuperscript{*}Note: Sub-titles are not captured in Xplore and should not be used}\thanks{Identify applicable funding agency here. If none, delete this.}
%\title{Multi-scale Pre-training with Perturbed Report Discrimination for Improved Biomedical Multi-modal Representation Learning}
\title{Enhancing Biomedical Multi-modal  Representation Learning with Multi-scale Pre-training and Perturbed Report Discrimination}
%\author{\IEEEauthorblockN{Anonymous Authors}}
%\iffalse
\author{
\IEEEauthorblockN{Xinliu Zhong}
\IEEEauthorblockA{\textit{Department of Biomedical Engineering} \\
\textit{National University of Singapore}\\
Singapore\\
xinliuzhong@u.nus.edu}
\and
\IEEEauthorblockN{Kayhan Batmanghelich}
\IEEEauthorblockA{\textit{Department of Electrical and Computer Engineering} \\
\textit{Boston University}\\
Boston, USA \\
batman@bu.edu}
\and
\IEEEauthorblockN{Li Sun}
\IEEEauthorblockA{\textit{Department of Electrical and Computer Engineering} \\
\textit{Boston University}\\
Boston, USA\\
lisun@bu.edu}

%\and
%\IEEEauthorblockN{4\textsuperscript{th} Wang Jia Hao}
%\and
%\and
%\IEEEauthorblockN{4\textsuperscript{th} Yeo Si Yong}
%\IEEEauthorblockA{\textit{Digital Health, Lee Kong Chian School of Medicine} \\
%\textit{Nanyang Technological University}\\
%Singapore \\
%siyong.yeo@ntu.edu.sg}
}
%\fi

\maketitle

\begin{abstract}
Vision-language models pre-trained on large scale of unlabeled biomedical images and associated reports learn generalizable semantic representations. These multi-modal representations can benefit various downstream tasks in the biomedical domain.
Contrastive learning is widely used to pre-train vision-language models for general natural images and associated captions. 
Despite its popularity, we found biomedical texts have complex and domain-specific semantics that are often neglected by common contrastive methods.
To address this issue, we propose a novel method, perturbed report discrimination, for pre-train biomedical vision-language models.
First, we curate a set of text perturbation methods that keep the same words, but disrupt the semantic structure of the sentence. Next, we apply different types of perturbation to reports, and use the model to distinguish the original report from the perturbed ones given the associated image.
Parallel to this, we enhance the sensitivity of our method to higher level of granularity for both modalities by contrasting attention-weighted image sub-regions and sub-words in the image-text pairs.
We conduct extensive experiments on multiple downstream tasks, and our method outperforms strong baseline methods. The results demonstrate that our approach learns more semantic meaningful and robust multi-modal representations.

\end{abstract}


\section{Introduction}
In the ever-evolving landscape of the biomedical field, the advent of Artificial Intelligence (AI) has unfurled a new era, particularly with its deep learning (DL) tendrils extending into diverse healthcare applications. DL's transformative role in medical imaging, embracing a spectrum of tasks from image classification to segmentation, has been metaphorically hailed as `A third eye for doctors' \cite{fourcade_deep_2019}.
Natural language processing (NLP) techniques have also been applied in the medical domain and achieved great success in many tasks, including radiology natural language inference, medical question answering, and others
%However, it's the subtle yet profound influence of AI-powered natural language processing (NLP) that's emerging as a cornerstone in digital health. 
%The rise of advanced pretrained self-supervised large language models (LLMs) like GPT-4\cite{openai2023gpt4}, Llama-2\cite{touvron2023llama}, and PaLM 2\cite{anil2023palm} has shown promise in handling clinical corpora with robust knowledge transfer techniques. 
Beyond imagery and linguistics, AI's prowess also manifests in the nuanced handling of varied data streams, including audio, video, and complex high-dimensional data, underscoring its multifaceted role in healthcare.

While AI has revolutionized healthcare, its advances have predominantly focused on single modalities. However, the intrinsic nature of healthcare information necessitates exploring multi-modal data. Multi-modal learning (MML), employing tokenization and embedding approaches, treats inputs from any modality as a token sequence to learn a joint embedding space. This approach not only mitigates data scarcity but also enables the development of diverse tasks and facilitates zero-shot learning \cite{xian_zero-shot_2020} without specific fine-tuning.

Vision-language models (VLM) are one typical type of multi-modal models, with popular methods such as CLIP \cite{radford2021learning} and ALIGN \cite{jia2021scaling} being successful. Although showing good performance, they grapple with challenges in relational understanding, fusion, and alignment between text and image modalities. VLMs also struggle with limited transferability, efficiency, universality, and interpretability. Within healthcare, these challenges are compounded by growing reporting backlogs, pressure on clinicians, such as strict radiology workflows slowing down case throughput especially as patient volumes rise, the absence of pretrained foundational models impeding rapid integration and deployment, and limited size of clinical datasets restricting the development of robust and versatile AI solutions. A particular hurdle is the unique healthcare discourse style and textual semantics, like the use of negation to rule out conditions (e.g., `no evidence of pneumonia'), which could be ignored by AI models without sufficient data fed in to interpret complex medical data such as paired chest X-rays (CXR) and corresponding radiological reports written by experienced physicians.

Aware of the deficiency of clinical semantics understanding of current language models as well as vision-language models, \cite{boecking_making_2022} developed CXR-BERT tailored specifically in outstanding CXR reports by additional pre-training steps focused on radiology reports to capture their dense and complex semantics. While this specialization significantly enhanced vocabulary-specific understanding, it did not fully address the intricacies of clinical sentence structure. Together with the specialized text model, they presented a self-supervised vision-language approach BioViL using contrastive learning, demonstrating state-of-the-art performance in various biomedical benchmarks.  

In this paper, we aim to propose a novel self-supervised VLM based on contrastive learning of jointly pretrained language and image models tailored for fine-grained understanding of clinical semantics and sentence structure, resulting in stronger vision-language alignment. Regarding high level of granularity of medical modalities, for example, the model should learn localized representations, where scene-level global information does not suffice. This is solved by introducing a local attentive contrastive loss which contrasts attention-weighted image sub-regions and words in the Image-Text pairs. Besides, the model should avoid the `object bias' \cite{yuksekgonul_when_2022} similar to `bag of objects' caused by CLIP-loss computation by our data-driven approach of loss design. We designed a set of diverse perturbations for every text in the original dataset, and the loss is computed reflecting the model's ability to discriminate these generated texts together with the original text, which share almost or totally same retrieval results. To examine whether our proposed model has better compositional representations of objects, attributes, and relations for medical text encoders, we tailored an evaluation protocol using the aforementioned text perturbations. The main contributions of this paper can be summarized as follows:
\begin{enumerate}
    \item We develop a novel self-supervised vision-language model that manipulates clinical texture well using diverse text perturbations derived from the standard image-text paired dataset in contrastive learning paradigm.
    \item We introduce a hierarchical contrastive learning strategy, including a image-level contrastive loss to capture global structure, and a local contrastive loss which focus on fine details.
    \item We further design a set of evaluation tasks to evaluate the model's ability to understand clinical language representation other than just `bag of words' based on the former perturbations.
    \item Empirically, our proposed model outperforms the baseline vision-language model on our proposed evaluation protocol of sentence structure, medical natural language inference benchmarks including MedNLI \cite{romanov2018lessons} and RadNLI \cite{miura-etal-2021-improving}. Especially, our model succeeds in multi-task image classification on a chest radiograph benchmark CheXpert \cite{irvin_chexpert_2019},  and outperforms baseline metods, demonstrating its strong ability in cross-modal correlation.
\end{enumerate}

\section{Related work }Our work lies in the evolving field of biomedical VLMs, which centers on extracting information from combination of medical images and reports. This area has recently gained traction, as evidenced by works like \cite{boecking_making_2022}. The majority of current methods rely on contrastive learning to bridge visual and linguistic elements. ConVIRT \cite{zhang2022contrastive} firstly used bidirectional contrastive losses to project the two modalities into a shared space, and performed image classification task well while requiring much fewer annotated training data as an ImageNet-initialized counterpart. A shift in focus towards local rather than global information was introduced by LoVT \cite{M_ller_2022}, which targets localized medical imaging tasks like semantic segmentation or object detection by introducing a local contrastive loss to align image regions or report sentences while encouraging spatial smoothness and sensitivity. Wang et al. \cite{wang2022medclip} extracted both image labels and entities extracted from reports to deepen understanding of medical semantics, eliminating the false negative noises and making use of vast image-only and text-only unpaired datasets. With the help of strong backbones, models such as BioViL \cite{boecking_making_2022} and ChexZero \cite{tiu2022expert}, which leverage a pretrained radiology-specific CXR-BERT text encoder and a pretrained CLIP \cite{radford2021learning} model respectively, also achieve comparable results.

Inspired by these prior studies, our work aims to comprehensively investigate the distinct clinical semantics as well as the high granularity of both medical vision and language that are essential to understanding biomedical multi-modal representation. 

\section{Method}
\begin{figure*}[htbp]
\centerline{\includegraphics[width=0.8\textwidth]{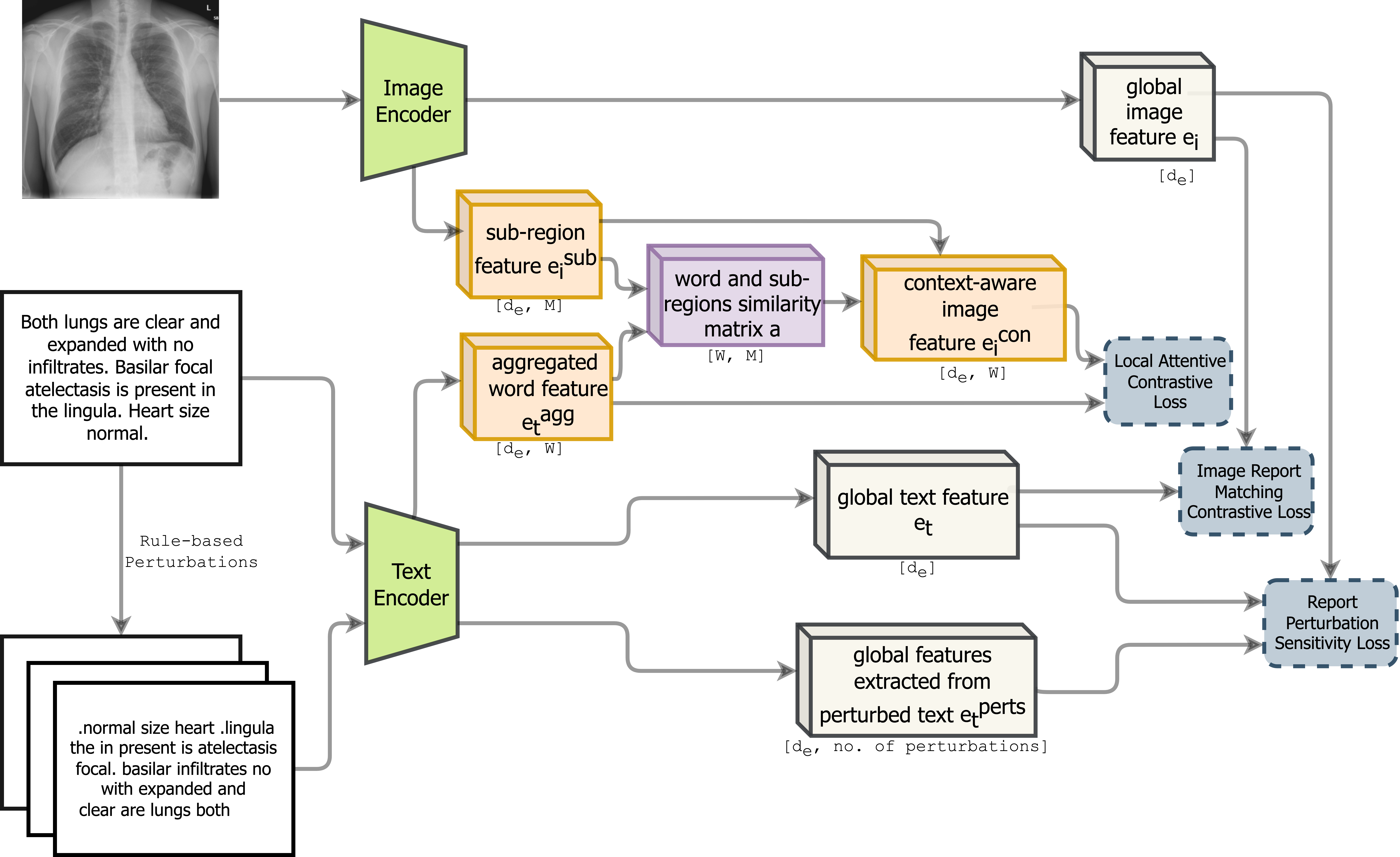}}
\caption{Overall architecture of our proposed self-supervised vision-language learning framework.}
\label{fig:per}
\end{figure*}
Our proposed self-supervised vision-language learning architecture integrates unimodal feature extractors for both images and texts pretrained separately in advance and a contrastive projection approach to fuse the cross-modal embeddings into joint space. Our local-attentive contrastive loss is designed to boost precision in medical image-text matching by aligning specific image sub-regions with relevant text fragments. We also introduce report perturbation sensitivity loss to understand clinical semantics, focusing on sentence structure and part of speech relevant to clinical contexts, paired with the standard image report matching contrastive loss which primarily focuses on lexical identification. A series of text perturbations, integral to this process, are also employed in our subsequent evaluation protocol. The overall architecture is shown in Fig. ~\ref{fig:per}.

\subsection{Pretrained Unimodal Feature Extractor}
\label{sub:fe}
With pre-trained text feature extractors carrying radiological knowledge by pretraining at hand, we choose to transfer powerful domain-specific language information from them instead of training the models from scratch which is time and energy-consuming. For each modality, we obtain the global features for subsequent calculation of image report matching contrastive loss and report perturbation sensitivity loss, as well as the local features for local attentive contrastive loss.
\iffalse
\subsubsection{Pretrained Vision Feature Extractor}
For image data, we used a ResNet-50 \cite{he2015deep} model pretrained on MIMIC-CXR \cite{johnson_mimic-cxr_nodate} images using SimCLR \cite{pmlr-v119-chen20j} with domain-specific augmentations including random affine transformations, random color jitter, and horizontal flips. Denoting the global embedding dimension as \(d_e\), the global features of the image \(e_i\in \mathbb{R}^{d_e}\) are extracted from the last layer. For \(M\) sub-regions in an image, every local embedding of the sub-region \(e^{local}_i\in [d_e, M]\) is extracted from an intermediate layer given.
\fi
%\subsubsection{Pretrained Language Feature Extractor}
For text data, we used the CXR-BERT pretrained via Masked Language Modelling (MLM) on the PubMed \cite{pubmed}, MIMIC-III \cite{johnson_mimic-iii_2016} and MIMIC-CXR corpora as well as radiology section matching (RSM) which sequence prediction tasks were performed on MIMIC-CXR to match `Impression' to `Findings' sections of the same study. The output of the pretrained model become the global features of the text \(e_t\in \mathbb{R}^{d_e}\). Considering the common abbreviations and typographical errors in medical reports, we employ word-piece level tokenization, resulting a sub-word feature \(e^{sub}_t\in [K, N]\) for every report with \(W\) words, \(K\) dimension for every word-piece feature, and \(N\) as the sum of the number of the sub-words in each word, output from the encoder. The ultimate word-level local text embedding \(e^{local}_t\in \mathbb{R}^{d_e}\) is aggregated by the average of the sub-word embeddings for each word, projected to a vector sized of \(d_e\).

\subsection{Joint Contrastive Projection}
Based on the gained unimodal information in radiology domain from aforementioned feature extractors, we build our own contrastive projection model to fuse the vision and language embeddings into the joint space. Target at tackling the issue of high granularity of medical region of interests and unique clinical texture and semantics style which differs from the general language, we incorporated the local attentive contrastive loss to learn the salient image sub-region to attend to, report perturbation sensitive loss to learn the linguistic characteristics as well as the image report matching contrastive loss to perform the simple lexical retrieval task. 

\subsubsection{Image Report Matching Contrastive Loss}
The basic image report matching contrastive loss \cite{radford2021learning} can be defined as a symmetric cross-entropy loss based on cosine similarity between image and text features. Denoting the normalized image and text global embeddings as \(e_i\in \mathbb{R}^{d_e}\) and \(e_t\in \mathbb{R}^{d_e}\) correspondingly, cross-entropy loss for Image-Text and Text-Image of the scaled pairwise similarity for every batch \(B\) can be defined as below,
%\begin{equation}
%    logits = E_t\times E_i^\intercal * e^t
%\end{equation}
where \(\tau\) is the scaling parameter:
%Addition of the symmetrical cross entropy loss for Image-Text and Text-Image becomes the final contrastive loss:
%\begin{equation}
%    labels = [0, 1, ..., n]
%\end{equation}
%\begin{equation}
%    loss = -\frac{log\frac{exp(logits, labels)}{\sum_{i=0}^n exp(logits_i)}+log\frac{exp(logits^\intercal, labels)}{\sum_{i=0}^n exp(logits^\intercal_i)}}{2}
%\end{equation}
\begin{align*}
\mathcal{L}_{global}=&-\log\frac{\exp(e_{i_j}^T\cdot e_{t_j}/\tau)}{\exp(e_{i_j}^T\cdot e_{t_j}/\tau)+\sum^{k\neq j} \exp(e_{i_k}^T\cdot e_{t_j}/\tau)}
%&-\log\frac{\exp(e_{i_j}^T\cdot e_{t_j}/\tau)}{\exp(e_{i_j}^T\cdot e_{t_j}/\tau)+\sum^{k\neq j} \exp(e_{i_j}^T\cdot e_{t_j}/\tau)}
\end{align*}
By employing this image report matching loss, the matched Image-Text pair were attracted together, while unmatched pairs were pushed away.
\subsubsection{Local Attentive Contrastive Loss}
Medical images' Region of Interests (ROI) are identified by more subtle cues than the natural images. This means only a small portion in a single graph aligns with the key word in the given report, requiring learning of local granules in addition to global vectors for Image-Text retrieval. In our local attentive contrastive loss \cite{huang2021gloria}, We adopted a context-aware image embedding \(e^{con}_i\in [d_e, W]\) for every image, which is the attention-weighted sum of all \(M\) sub-region’s significance \(e^{sig}_i\in [M, W]\) to every given word. The attention weight \(a_{jk}\) is the normalized similarity between a given word \(j\) and a given sub-region \(k\) across all image sub-regions: 

\begin{align*}
    a_{jk} = \frac{\exp{({{{e^{local}_{i_k}}}^T}\cdot e^{local}_{t_j})}}{\sum^{M}_{l}\exp{({{{e^{local}_{i_l}}}^T}\cdot e^{local}_{t_j})}} 
\end{align*}

The aggregated similarities \(S\) between all word embeddings and their corresponding context-aware image embeddings defines the local attentive matching score for a Image-Text pair \([I, T]\):
\begin{align*}
    \mathcal{S_{I,T}}=\log(\sum^W_{k=1}{(\exp({\sum^M_{j=1}{a_{jk}\cdot v_k}}))^T\cdot{e^{local}_{t_k}}})
\end{align*}
Addition of the symmetrical cross-entropy loss for Image-Text and Text-Image matching becomes the final contrastive loss:
\begin{align*}
\mathcal{L}_{local} = -\log\frac{\exp({S_{I_i,T_i}/\tau)}}{{\exp(S_{I_i,T_i}/\tau)+\sum^{j\neq i} }\exp({S_{I_i,T_j}/\tau)}}
\end{align*}
The model learns the fine-grained local semantic alignments between the image sub-regions and word pieces by employing this attentive contrastive loss. 

\subsubsection{Report Perturbation Sensitivity Loss}
We present a report perturbation sensitivity loss to teach our model to understand the rich semantics in the radiological setting. Without expensive manual annotations or specific encoding of order and composition cues in advance, our method can enhance the text encoder's preference for reports with correct word ordering and semantics. This is done by performing perturbations of every caption systematically based on our customized rules using spaCy \cite{spacy2}. 
\paragraph{Generating Text Perturbations}
In \cite{oconnor-andreas-2021-context}, the author conducted a series of ablation experiments on language models, demonstrating several manipulations of the contexts are crucial in degrading the models' performance. Inspired by that, we define a set of rules and generate 9 different perturbations as negatives by shuffling the words of specific entity types or according to specific patterns while keeping words unchanged. These perturbations, shown in Table \ref{tab:per}, are used both in the calculation of report perturbation sensitivity loss and subsequent evaluation.
\begin{table*}[!htbp]
    \caption{List of text perturbations for an example radiological report}
    \label{tab:per}
    \centering
    %\tiny
    %\begin{tabularx}{0.45\textwidth}{p{0.15\textwidth}p{0.3\textwidth}} 
    \begin{tabular}{ll}
         \toprule
         \textbf{Perturbation Type}& \textbf{Example}\\
         \midrule
         Original Text& the lungs are clear there is no pleural effusion or pneumothorax\\
         Shuffle All Words& pneumothorax are lungs the there is pleural or no effusion clear
\\
 Swap Adjacent Words in the Sentence&lungs the clear are is there pleural no or effusion pneumothorax\\
         Reverse Sentence& pneumothorax or effusion pleural no is there clear are lungs the\\
         Shuffle Within Trigrams& lungs the are there is clear pleural effusion no or pneumothorax\\
         Shuffle Trigrams& or pneumothorax no pleural effusion the lungs are clear there is\\
         Shuffle Nouns and Adjectives& the pneumothorax are clear there is no pleural lungs or effusion\\
         Shuffle All but Nouns and Adjectives& there lungs is clear are the or pleural effusion no pneumothorax\\
         Shuffle Nouns, Verbs, and Adjectives& the is pneumothorax lungs there pleural no are effusion or clear\\
         Replace Adjectives with Antonyms& the lungs are unclear there is no pleural effusion or pneumothorax\\
         \bottomrule
    %\end{tabularx}
    \end{tabular}
\end{table*}
\paragraph{Contrastive Loss}
Deriving from the original captions, we use the generated perturbations as negative text samples in contrast with the original reports. The global image embeddings are denoted \(e_{i_{ori}}\in \mathbb{R}^{d_e}\), and the global text embeddings from the original texts and the 9 generated perturbations are denoted as \(e_{t_{ori}}\in \mathbb{R}^{d_e}\) and \(e_{t_{Perts}} \in [d_e, 9]\). For a single image, the report perturbation contrastive loss is calculated as the cross-entropy loss of the similarity with the original aligned text against similarity with all other negatives:

\begin{align*}
\mathcal{L}_{pert}=&-\log\frac{\exp(e_{i_{ori}}^T\cdot e_{t_{ori}}/\tau)}{\exp(e_{i_{ori}}^T\cdot e_{t_{ori}}/\tau)+\sum^{Perts}_{k} \exp(e_{i_{ori}}^T\cdot e_{t_{k}}/\tau)}
%&-\log\frac{\exp(e_{i_j}^T\cdot e_{t_j}/\tau)}{\exp(e_{i_j}^T\cdot e_{t_j}/\tau)+\sum^{k\neq j} \exp(e_{i_j}^T\cdot e_{t_j}/\tau)}
\end{align*}
This way, the model aims to pick the original reports for an associated image, thus understanding the clinical contexts better.

Therefore, our final loss is:
\begin{equation*}
\mathcal{L}=\mathcal{L}_{global} + \alpha \mathcal{L}_{local} + \beta \mathcal{L}_{pert}
\end{equation*}

\section{Experiment}
To validate our proposed model, we conducted various downstream evaluation experiments including fine-tuning clinical text classification tasks on RadNLI and MedNLI, fine-tuning multi-task image classification task on CheXpert, as well as our own clinical semantic structure evaluation task on Open-I dataset after pretraining.
\subsection{Datasets}
\subsubsection{Open-I \cite{openi}}
We used the Open-I dataset to train our vision-language model. The chest X-ray dataset contains 3,996 radiology reports associated with 8,121 images. The \textit{findings} part of the reports are used, and 6469 Image-Text pairs remains after removing empty values. 
%We extracted 6000 Image-Text pairs for model training, and 469 for clinical semantic structure evaluation.
\subsubsection{RadNLI \cite{miura-etal-2021-improving} and MedNLI \cite{romanov2018lessons}}
RadNLI and MedNLI benchmarks contain 960 and 11k labeled hypothesis and premise pairs, respectively. MedNLI is collected in a broad clinical domain, while RadNLI is more radiology-specific. The task, also named Natural Language Inference (NLI), is to predict the label from the three label categories: entailment, contradiction, and neutral given the sentence pair.

\subsubsection{CheXpert \cite{irvin_chexpert_2019}}
The CheXpert dataset contains a total of 224,316 chest radiographs from 65,240 patients, where each radiograph is paired with the corresponding radiology reports. Each radiograph is labeled for the presence of 14 total medical observations. 
%We only used frontal images, 20596 for training and 500 images for testing. Following the setting in \cite{huang2021gloria}, we chose Atelectasis, Cardiomegaly, Edema, Pleural, Effsion labels for multi-task image classification.

\subsection{Implementation details}
In all scenarios, we pretrained our multi-modal model for 150 epochs, with a batch size of 64. We use SGD \cite{sgd} optimizer, and set the learning rate to 0.0015, momentum factor to 0.9, and weight decay to \(5e-4\). For image preprocessing, we follow \cite{boecking_making_2022}, using the same augmentations as used in the aforementioned pretrained image encoder. The dimensions of our input images are  \(224 \times 224\) pixels. To fine-tune the impact of each loss term and the margin in our model, we set the scaling hyper-parameters \(\alpha\) and \(\beta\) at 0.1, and \(\tau\) at 0.07. These values were selected based on outcomes from a targeted grid search.

\subsection{Evaluation of extracted representation}
\begin{table*}[htp]
\caption{Results of fine-tuned multi-task image classification on the CheXpert benchmark}
\begin{center}
%\setlength\tabcolsep{0pt}
%\begin{tabular}{0.5\textwidth}{@{\extracolsep{\fill}} c|c|c|c|c|c|c}
\begin{tabular}{l c c  c}
\toprule
%\multirow{2}{4em}{\textbf{Method}} & \multicolumn{6}{c}{\textbf{CheXpert}} \\
%\cmidrule{2-7}
%\hline

Accuracy &  Consolidation  & Pleural Effusion & Average  \\
\midrule
ConVIRT \cite{zhang2022contrastive} &  28.80\%  & 43.60\% & 36.20\%\\
GLoRIA \cite{huang2021gloria} &   71.11\% & 28.89\% & 50.00\% \\
Ours (w/o local loss)               &  33.80\% &  \textbf{77.80\%} & 55.80\%\\
Ours                           &  \textbf{93.40\%} &  51.80\% & \textbf{72.60\%}\\
\bottomrule
%\multicolumn{3}{l}{$^{\mathrm{a}}$Sample of a Table footnote.}
\end{tabular}
\label{tab:image}
\end{center}
\end{table*}
We compare our model to its version that removes the local attentive contrastive loss, as well as two strong baselines under several configurations. 
ConVIRT \cite{zhang2022contrastive} (Contrastive VIsual Representation
Learning from Text) is an efficient method of image representation learning from natural language supervision by contrasting only global representations of image and report pairs. 
In addition, within the same radiological domain, we also compared our architecture to GLoRIA \cite{huang2021gloria} (Global-Local Representations for Images using Attention). This method also demonstrated high-performance and label-efficiency for various downstream medical image recognition tasks with limited labels. 
For a fair comparison, we keep the pretraining procedure unchanged, using the same 6000 Image-Text pairs of the Open-I dataset as input. The pretrained unimodal encoders are kept the same, using the same published checkpoints through the comparison. We conducted the same protocols on our downstream task as well. 

\subsubsection{Text classification}
For text classification, we fed the MedNLI and RadNLI datasets into the text encoder of the multi-modal model after pretraining and fine-tuning separately. Table \ref{tab:text} displays the accuracy of the different models on MedNLI and RadNLI benchmarks.
\begin{table}[htbp]
\caption{Results of fine-tuned text classification on the RadNLI and MedNLI benchmarks}
\begin{center}
\begin{tabular}{l c c}
\toprule
\textbf{Method} & \textbf{MedNLI} & \textbf{RadNLI} \\
\midrule
 ConVIRT \cite{zhang2022contrastive} & 86.80\% & 68.50\% \\
GLoRIA \cite{huang2021gloria}   & 86.64\% & 68.33\%\\
Ours (w/o local loss)               & \textbf{87.62\%} & 66.67\% \\
Ours                           & 85.79\% & \textbf{68.96\%} \\
\bottomrule
%\multicolumn{3}{l}{$^{\mathrm{a}}$Sample of a Table footnote.}
\end{tabular}
\label{tab:text}
\end{center}
\end{table}

\subsubsection{Multi-task image classification}
For fine-tuned multi-task image classification, we only used frontal images, and we use 20,596 images for training and 500 images for testing. We benchmark the performance on classification of \textit{Consoildation and  Pleural Effusion}. Similarly, we fed the dataset into the image encoder of the multi-modal model after pretraining and fine-tuning. Table \ref{tab:image} displays the accuracy of the different models on both separate tasks and averages.
\iffalse
\begin{table*}[htbp]
\caption{Results of fine-tuned multi-task image classification on the CheXpert benchmark}
\begin{center}
%\setlength\tabcolsep{0pt}
%\begin{tabular}{0.5\textwidth}{@{\extracolsep{\fill}} c|c|c|c|c|c|c}
\begin{tabular}{c c c c c c c}
\toprule
\multirow{2}{4em}{\textbf{Method}} & \multicolumn{6}{c}{\textbf{CheXpert}} \\
\cmidrule{2-7}
%\hline

 & Cardiomegaly & Edema & Consolidation  & Atelectasis  & Pleural Effusion & Mean  \\
\midrule
ConVIRT \cite{zhang2022contrastive} & 29.00\% & 68.40\% & 28.80\% & 58.00\% & 43.60\% & 48.02\%\\
GLoRIA \cite{huang2021gloria}   & 32.22\% & 74.44\% & 71.11\% & 64.44\% & 28.89\% & 43.67\% \\
Ours (w/o local)               & 24.00\% & 65.40\% & 33.80\% & 22.40\% & 77.80\% & 63.35\%\\
Ours                           & 28.20\% & 34.00\% & 93.40\% & 45.80\% & 51.80\% & 48.16\%\\
\bottomrule
%\multicolumn{3}{l}{$^{\mathrm{a}}$Sample of a Table footnote.}
\end{tabular}
\label{tab:image}
\end{center}
\end{table*}
\fi

%\subsubsection{Zero-shot clinical semantic structure evaluation}
\subsubsection{Clinical Semantic Structure Evaluation}
In our clinical semantic structure evaluation task, we used the aforementioned text perturbations as negatives. The cosine similarity is calculated for every caption inside a candidate caption list which includes an original text and its perturbations with the corresponding image using the global embeddings:
\begin{align*}
    \mathcal{S_{I,T}}=\frac{e_{I}\cdot e_{T}}{\left\lVert e_{I} \right\rVert_{2} \left\lVert  e_{T}\right\rVert_{2}}
\end{align*}

If the similarity between the original aligned Image-Text pair is the highest, then the decision is considered as correct. This means the model is able to retrieve the correctly structured report given an image. This evaluation task incorporates the representations of negative samples, outperforming the classical Image-Text retrieval task by introducing more difficult candidates.

We use 469 image-text pairs from the Open-I dataset as validation set, then feeding them into the uni-modal encoders of the multi-modal model, deriving the accuracy illustrated before. The results are shown in Table~\ref{tab:eva}. The results show that our proposed method outperforms baseline methods.

\begin{table}[htbp]
\caption{Results of zero-shot clinical semantic structure evaluation on the Open-I benchmark}
\begin{center}
\begin{tabular}{l c}
\toprule
\textbf{Method} &\textbf{ Open-I} \\
\midrule
ConVIRT \cite{zhang2022contrastive} & 43.10\% \\
GLoRIA \cite{huang2021gloria}   & 44.30\% \\
Ours (w/o local loss)               & 46.30\% \\
Ours                  & \textbf{49.00\%}\\
\bottomrule
%\multicolumn{3}{l}{$^{\mathrm{a}}$Sample of a Table footnote.}
\end{tabular}
\label{tab:eva}
\end{center}
\end{table}

%\section{Ablation}
%\input{chapters/Ablation}

\section{Conclusion}

In this paper, we propose a novel pre-trained vision-Language model that enhances fine-grained clinical semantics understanding by increasing sensitivity to caption perturbations and focusing on local attention. This approach allows the model to capture not just broad image-level information, but also intricate clinical sub-region and word-piece details within both images and texts. By contrasting generated negative examples with original medical reports, our model gains a deeper understanding of medical report semantics and structure. This method has demonstrated significant improvements across various clinical datasets in multiple downstream benchmarks and our custom-designed composition sensitivity evaluation task. %This task could potentially serve as a benchmark for evaluating the contextual understanding capabilities of Vision-Language Models (VLMs).
In the future, we will explore pre-training our proposed model with images of higher resolution, and extending our model to other modalities, including electronic health records (EHR). 

%\section*{References}

\printbibliography %Prints bibliography

\end{document}